\documentclass[10pt,twocolumn,letterpaper]{article}

\usepackage{iccv}              %

\definecolor{iccvblue}{rgb}{0.21,0.49,0.74}
\definecolor{kleinblue}{RGB}{0, 47, 167}

\usepackage[pagebackref,breaklinks,colorlinks,allcolors=kleinblue]{hyperref}

\def\paperID{5695} %
\def\confName{ICCV}
\def\confYear{2025}

\title{DiffCLIP: Differential Attention Meets CLIP}

\author{
Hasan Abed Al Kader Hammoud\thanks{Correspondence at\\\texttt{hasanabedalkader.hammoud@kaust.edu.sa}}\\
KAUST\\
\and
Bernard Ghanem\\
KAUST\\
}

\usepackage{booktabs}
\usepackage{multirow}
\usepackage{xcolor} %
\usepackage{array}  %

\definecolor{mygreen}{rgb}{0.0, 0.5, 0.0}   %
\definecolor{myred}{rgb}{0.6, 0.0, 0.0}     %

\newcommand{\posdiff}[1]{\textcolor{mygreen}{\scriptsize (+#1)}}
\newcommand{\negdiff}[1]{\textcolor{myred}{\scriptsize (-#1)}}

\usepackage{tcolorbox}

\newtcolorbox{conclusionbox}[1][]{
    colback=gray!10,           %
    colframe=black,            %
    boxrule=1pt,               %
    arc=0pt,                   %
    left=5mm,                  %
    right=5mm,                 %
    top=3mm,                   %
    bottom=3mm,                %
    before skip=15pt,          %
    after skip=15pt,           %
    halign=center,             %
    #1                         %
}

\definecolor{DarkRoyalBlue}{RGB}{52,84,180}

\usepackage{xcolor}

\begin{document}
\maketitle

\begin{abstract}
We propose DiffCLIP, a novel vision-language model that extends the differential attention mechanism to CLIP architectures. Differential attention was originally developed for large language models to amplify relevant context while canceling out noisy information. In this work, we integrate this mechanism into CLIP's dual encoder (image and text) framework. With minimal additional parameters, DiffCLIP achieves superior performance on image-text understanding tasks. Across zero-shot classification, retrieval, and robustness benchmarks, DiffCLIP consistently outperforms baseline CLIP models. Notably, these gains come with negligible computational overhead, demonstrating that differential attention can significantly enhance multi-modal representations without sacrificing efficiency. Code can be found at \url{https://github.com/hammoudhasan/DiffCLIP}.

\end{abstract}    
\section{Introduction}

Vision-language models (VLMs) have made remarkable progress in bridging the gap between textual and visual modalities, enabling powerful capabilities such as zero-shot image classification, image-text retrieval, and descriptive captioning \cite{radford2021learning, jia2021scaling}. By aligning images and text in a joint embedding space, these models capture broad semantic relationships across modalities and often excel at out-of-distribution generalization. Among VLMs, Contrastive Language-Image Pre-training (CLIP) \cite{radford2021learning} stands out as a foundational approach, demonstrating strong zero-shot performance on numerous benchmarks with minimal fine-tuning.

While CLIP’s contrastive training regime has been widely adopted, its \emph{attention mechanism} can sometimes focus on irrelevant or spurious features in both the image and text encoders. This \emph{attention noise} can hamper fine-grained understanding, particularly when precise localization or explicit contextual knowledge is required. Interestingly, recent language modeling research has proposed a \emph{differential attention} mechanism \cite{ye2024differential}, which subtracts complementary attention distributions to suppress noise and highlight salient tokens. However, whether a similar strategy would be effective for multimodal tasks has remained an open question.

\begin{center}
    \textcolor{DarkRoyalBlue}{\emph{``\textbf{Can differential attention be adapted to vision-language models in a way that meaningfully improves their ability to focus on relevant features across modalities?}''}}
\end{center}

\begin{figure}[t]
    \centering
    \includegraphics[width=\linewidth]{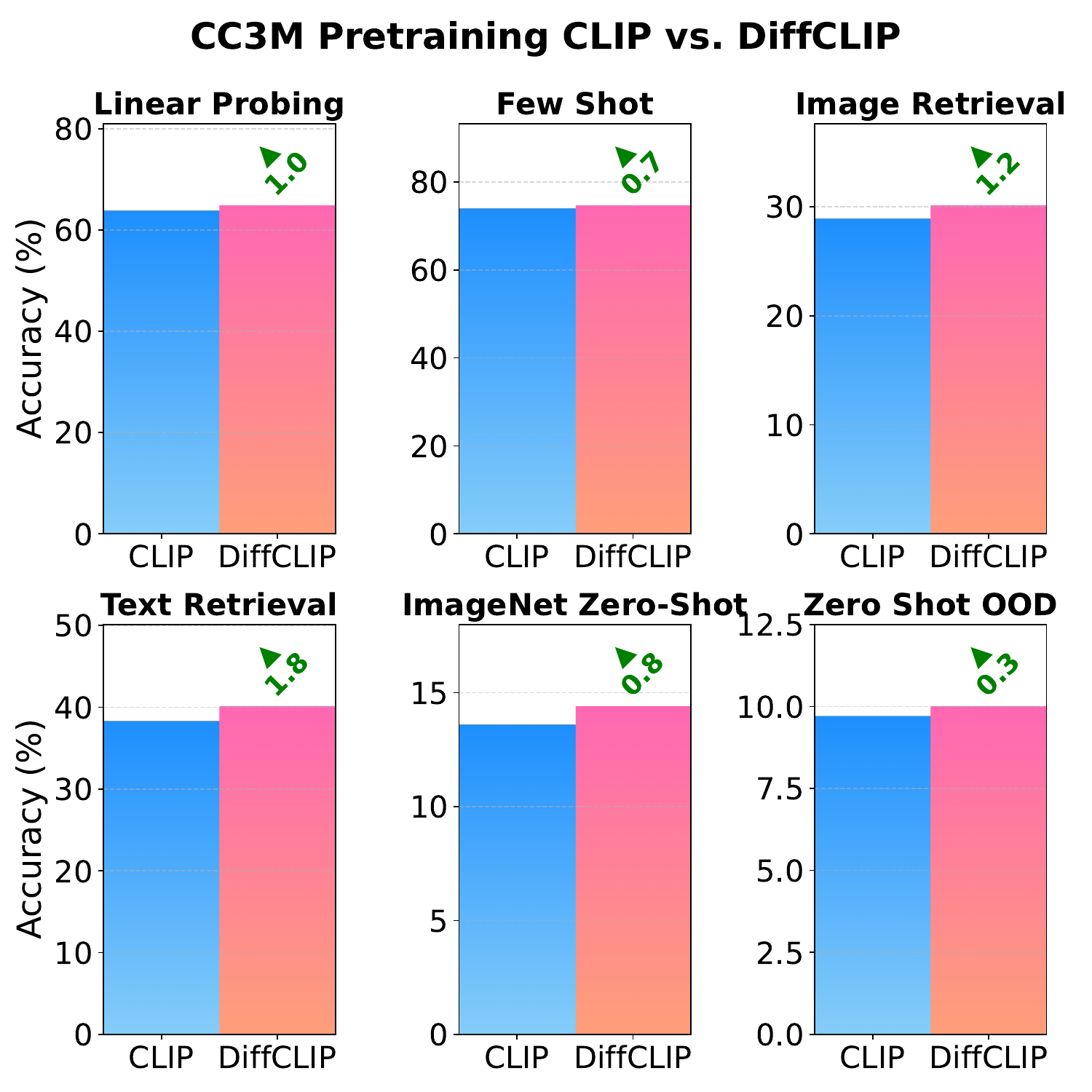}
\caption{\textbf{CC3M Pretraining: CLIP vs.\ DiffCLIP Across Six Tasks.}
We compare standard CLIP (blue) and our DiffCLIP variant (pink) on linear probing, few-shot classification, image/text retrieval, zero-shot ImageNet, and zero-shot OOD. 
In each case, DiffCLIP consistently outperforms CLIP, highlighting the effectiveness of differential attention with only 0.003\% extra parameters.}
    \label{fig:pull_figure}
\end{figure}

\begin{figure*}
    \centering
    \includegraphics[width=\linewidth]{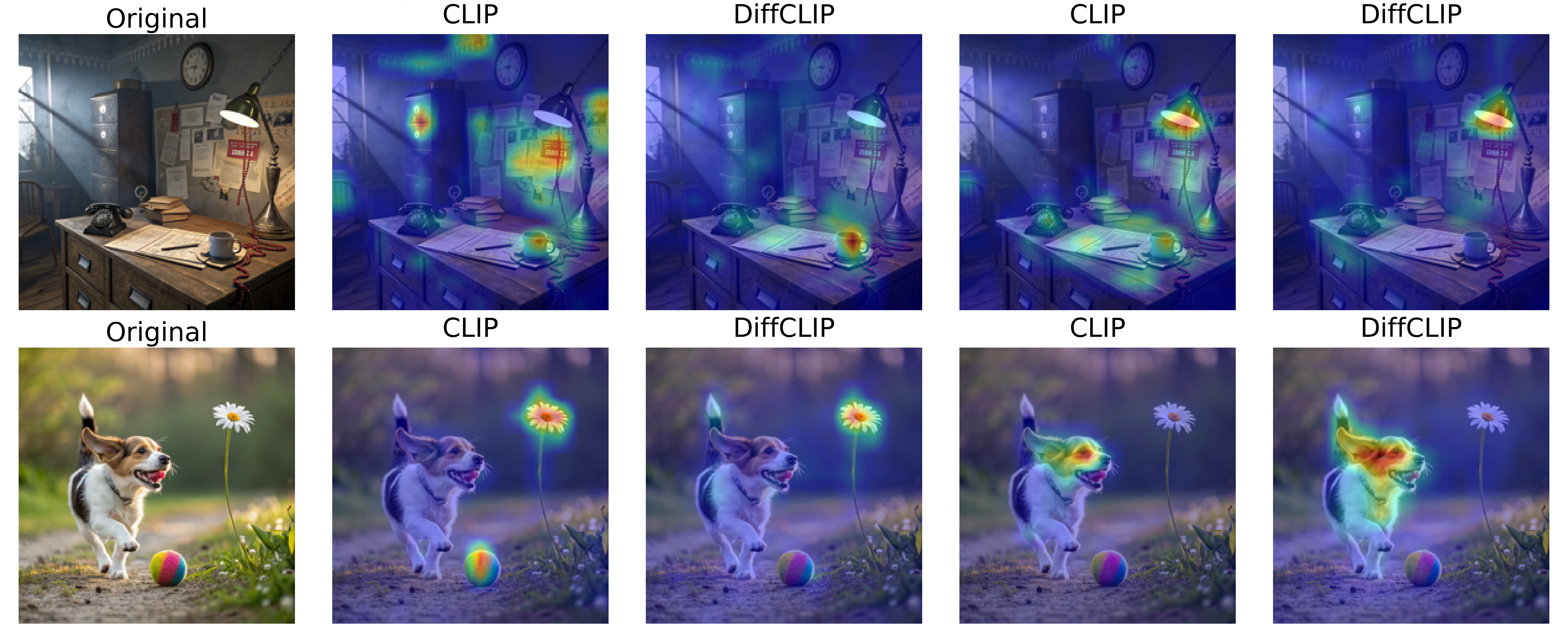}
\caption{\textbf{Comparing CLIP vs.\ DiffCLIP Attention Maps.}
For two images (rows), we visualize where CLIP and DiffCLIP attend when matching each image against two different textual queries. 
While CLIP allocates attention to irrelevant background regions, DiffCLIP more effectively centers on query-relevant objects, highlighting how differential attention can reduce noise and improve focus. \\ \textcolor{DarkRoyalBlue}{\textit{Queries: \textbf{First Row:} `Mug'', Lamp''; \textbf{Second Row:} Flower'', Dog''.}}}
\label{fig:attention_maps}
\end{figure*}

Motivated by this question, we introduce {DiffCLIP}, an extension of CLIP that integrates differential attention into both the vision and text encoders. By learning two attention maps and subtracting one from the other, DiffCLIP effectively cancels out misaligned or noisy signals, enabling a more precise alignment of images and text. Crucially, this enhancement introduces only a negligible overhead in model parameters and computational cost. Our results show that DiffCLIP consistently outperforms standard CLIP on a wide range of tasks—including linear probing, few-shot classification, image-text retrieval, out-of-domain robustness, and fine-grained visual understanding—highlighting the efficacy of differential attention in a multimodal setting. As shown in Figure \ref{fig:pull_figure}, DiffCLIP is capable of improving performance across various benchmarks with only $0.003\%$ extra parameters. Figure \ref{fig:attention_maps} also shows how DiffCLIP is capable of suppressing attention noise compared to CLIP models with vanilla non-differential attention.

Our contributions are threefold:
\begin{itemize}
    \item We propose {DiffCLIP}, the first integration of differential attention into CLIP-based VLMs, yielding a simple yet effective approach to reducing attention noise in both vision and text streams.
    \item Through extensive experiments on Conceptual Captions 3M/12M pretraining, we demonstrate consistent gains over baseline CLIP across a diverse suite of tasks, with a minimal parameter overhead of roughly $0.003\%$.
    \item We perform detailed ablations, showing that \emph{(i)}~dynamic initialization can boost zero-shot performance, and \emph{(ii)}~applying differential attention solely in the vision encoder already captures most of the benefits, suggesting a flexible, cost-effective path to improved multimodal learning.
\end{itemize}

The remainder of this paper is organized as follows. Section~\ref{sec:related} surveys previous work on training-centric, model-centric, and data-centric strategies for enhancing CLIP. Section~\ref{sec:preliminaries} provides an overview of the standard Transformer attention mechanism, the differential attention concept, and the CLIP framework. Section~\ref{sec:experiments} details our experimental setup, empirical results, and ablation studies, while Sections~\ref{sec:future} and \ref{sec:conclusion} concludes with a discussion of future research directions and wrapping up of the paper.

\section{Related Work}
\label{sec:related}

Vision-language pre-training (VLP) has advanced our ability to learn joint representations of images and text, leading to improvements in tasks such as image retrieval, visual question answering, and zero-shot classification \cite{gan2022vision,zhang2024vision}. CLIP \cite{radford2021learning} has been central to this progress by using a contrastive loss to align image and text embeddings from large-scale image-caption data. Despite CLIP’s strong zero-shot performance, researchers continue to explore improvements in its training, architecture, and data collection strategies. These efforts generally fall into three categories: training-centric, model-centric, and data-centric approaches.

\paragraph{Training-Centric Approaches}
A common strategy is to enrich CLIP’s contrastive framework with additional objectives. For example, SLIP \cite{mu2022slip} adds masked image modeling to boost downstream results, while DeCLIP \cite{li2021supervision} uses nearest-neighbor supervision to enhance data efficiency. SigLIP \cite{zhai2023sigmoid} replaces the standard softmax temperature with a sigmoid loss, allowing larger batch training and improving generalization and robustness to noisy labels. Retrieval-Enhanced CLIP \cite{iscen2023retrieval} leverages external memory of image-text pairs at inference, achieving significant gains on fine-grained zero-shot tasks. Further, novel training objectives, such as those proposed by Yang et al. \cite{yang2022unified} and PyramidCLIP \cite{gao2022pyramidclip}, aggregate information across multiple semantic levels, highlighting the benefit of diversified training signals for improved CLIP performance.

\paragraph{Model-Centric Approaches}
Another line of work modifies CLIP’s architecture for greater efficiency or accuracy. The original CLIP \cite{radford2021learning} employs a Transformer \cite{vaswani2017attention} for text and either a ResNet \cite{he2016deep} or Vision Transformer (ViT) \cite{dosovitskiy2020image} for images. Subsequent studies incorporate ideas from object detection and segmentation to capture finer visual details, such as region-level representations \cite{xu2022groupvit, zhong2022regionclip}. Recently, ViTamin \cite{chen2024vitamin} proposed a specialized vision transformer architecture tailored specifically for multimodal models, demonstrating improved zero-shot results compared to standard ViTs under similar training setups. Other researchers attempt to unify image and text encoders into a single Transformer \cite{tschannen2022image}, although this approach is less common. Notably, few methods have altered the core attention mechanism within CLIP. Our work addresses this gap by adapting Differential Attention \cite{ye2024differential}, originally proposed for language models, to CLIP’s multimodal setting. This adaptation aims to reduce attention noise and enhance representation quality.

\paragraph{Data-Centric Approaches}
Data-centric methods emphasize improving the size, diversity, and quality of pre-training datasets. Initial efforts focused on scaling datasets \cite{jia2021scaling, radford2021learning}, while more recent approaches prioritize richer and cleaner supervision. VeCLIP \cite{lai2024veclip} uses large language models (LLMs) to generate detailed and enriched captions, enhancing textual supervision. Similarly, CLIPS \cite{liu2024clips} utilizes truncated synthetic captions to improve visual grounding and retrieval performance, showing that carefully controlled synthetic textual inputs can surpass standard image-caption pairs. SynthCLIP \cite{hammoud2024synthclip} explores training entirely on synthetic image-text pairs. Further methods employ filtering techniques to eliminate noisy or irrelevant samples \cite{gadre2023datacomp, abbas2023semdedup}, while Cluster Masking \cite{wei2024efficient} proposes masking clusters of similar image patches, leading to faster training and improved representation quality. These efforts underline the potential of data curation and augmentation strategies in bolstering the efficacy of CLIP-based models.\\

Beyond performance, fairness and compositionality have also received increased attention. FairCLIP \cite{luo2024fairclip} addresses demographic biases found in models like CLIP by using optimal-transport-based feature alignment across demographic groups. Meanwhile, iterated learning approaches \cite{zheng2024iterated} tackle the compositional limitations of large vision-language models, promoting representations that generalize more reliably to complex and compositional visual-linguistic scenarios.

\begin{tcolorbox}[width=\linewidth,colback=blue!2,colframe=blue!50!black,arc=2mm,boxrule=0pt]
In this paper, we contribute to the model-centric direction by adapting Differential Attention \cite{ye2024differential} to CLIP’s dual-encoder architecture. Through this adaptation, we aim to reduce attention noise and enhance performance across various image-text understanding tasks.
\end{tcolorbox}

\section{Preliminaries}
\label{sec:preliminaries}

In this section, we outline the fundamental concepts that are essential for our approach. We begin by reviewing the Transformer self-attention mechanism \cite{vaswani2017attention}, which is widely used in modern sequence modeling. Next, we introduce differential attention \cite{ye2024differential}, a technique designed to reduce attention noise by leveraging complementary attention distributions. Finally, we summarize the Contrastive Language-Image Pre-training (CLIP) framework \cite{radford2021learning}, which learns to align images and text in a shared representation space. These components form the basis for our model and experiments.

\subsection{Transformer Attention}
\label{subsec:transformer_attention}

Transformer networks \cite{vaswani2017attention} capture relationships among elements in a sequence through a \emph{self-attention} operation. Let
\[
X \in \mathbb{R}^{N \times d}
\]
be an input sequence of \(N\) tokens (or image patches), each embedded in a \(d\)-dimensional space. The Transformer maps \(X\) to queries (\(\mathbf{Q}\)), keys (\(\mathbf{K}\)), and values (\(\mathbf{V}\)) using learned weight matrices:
\[
\mathbf{Q} = X \, W^Q,
\quad
\mathbf{K} = X \, W^K,
\quad
\mathbf{V} = X \, W^V,
\]
where \(W^Q, W^K, W^V \in \mathbb{R}^{d \times d}\). Self-attention scores are then computed via scaled dot-products:
\[
A 
\;=\;
\mathrm{softmax}\!\Bigl(\tfrac{\mathbf{Q}\mathbf{K}^\top}{\sqrt{d}}\Bigr),
\]
and these scores are used to weight \(\mathbf{V}\):
\[
\mathrm{Attn}(X)
\;=\;
A \,\mathbf{V}.
\]

To capture different types of relationships, Transformers use \emph{multi-head attention} (MHA). An MHA module with \(h\) heads splits each projection into lower-dimensional parts of size \(d_h = d / h\). In each head \(i\),
\[
\mathrm{Attn}_i(X)
\;=\;
\mathrm{softmax}\!\Bigl(\tfrac{\mathbf{Q}_i \mathbf{K}_i^\top}{\sqrt{d_h}}\Bigr) \mathbf{V}_i,
\]
where \(\mathbf{Q}_i, \mathbf{K}_i, \mathbf{V}_i \in \mathbb{R}^{N \times d_h}\). The head outputs are concatenated and projected back:
\[
\mathrm{MHA}(X)
\;=\;
\bigl[\mathrm{Attn}_1(X)\,\|\,\dots\,\|\,\mathrm{Attn}_h(X)\bigr]\,
W^O,
\]
with \(W^O \in \mathbb{R}^{(h\,d_h)\times d}\). Despite remarkable success in many areas, standard attention can assign non-negligible weights to irrelevant tokens (often called \emph{attention noise}) \cite{needle,lost}, which can degrade performance in settings requiring precise focus.

\subsection{Differential Attention}
\label{subsec:differential_attention}

Differential attention \cite{ye2024differential} addresses attention noise by learning two separate attention distributions and subtracting one from the other, effectively canceling out spurious alignments.

\vspace{4pt}
\noindent\textbf{Single-Head Differential Attention.}\quad
Let
\(
X \in \mathbb{R}^{N \times d}
\)
be the input to a single attention head. We split \(\mathbf{Q}\) and \(\mathbf{K}\) into two halves, denoted by subscripts 1 and 2:
\[
[\mathbf{Q}_1;\,\mathbf{Q}_2] = X\,W^Q,
\quad
[\mathbf{K}_1;\,\mathbf{K}_2] = X\,W^K,
\quad
\mathbf{V} = X\,W^V,
\]
where \(\mathbf{Q}_1, \mathbf{Q}_2, \mathbf{K}_1, \mathbf{K}_2 \in \mathbb{R}^{N \times \frac{d}{2}}\). Each half computes its own attention distribution:
\[
A_1
=
\mathrm{softmax}\!\Bigl(\tfrac{\mathbf{Q}_1 \mathbf{K}_1^\top}{\sqrt{d/2}}\Bigr),
\quad
A_2
=
\mathrm{softmax}\!\Bigl(\tfrac{\mathbf{Q}_2 \mathbf{K}_2^\top}{\sqrt{d/2}}\Bigr).
\]
The output is formed by subtracting the second distribution (scaled by a learnable parameter \(\lambda\)) from the first:
\[
\mathrm{DiffAttn}(X)
=
\bigl(A_1 - \lambda\,A_2\bigr)\,\mathbf{V}.
\]
The parameter \(\lambda\) is trained to control how strongly the second distribution is subtracted:
\[
\lambda 
\;=\; 
\exp\!\bigl(\lambda_{q_1}\,\lambda_{k_1}\bigr)
\;-\; 
\exp\!\bigl(\lambda_{q_2}\,\lambda_{k_2}\bigr)
\;+\;
\lambda_{\mathrm{init}},
\]
where \(\lambda_{q_1}, \lambda_{k_1}, \lambda_{q_2}, \lambda_{k_2}\) are learnable weights and \(\lambda_{\mathrm{init}}\) is a hyperparameter. This subtraction often yields a sparser, more focused attention map, which can improve results in scenarios sensitive to background or redundant signals \cite{ye2024differential}.

\vspace{4pt}
\noindent\textbf{Multi-Head Extension.}\quad
Like standard attention, differential attention can be extended to multiple heads. In \emph{Differential Multi-Head Attention} (Diff MHA), each head \(i\) applies the differential step independently:
\[
\begin{aligned}
\mathrm{DiffAttn}_i(X)
&=
\Biggl(
\mathrm{softmax}\!\Bigl(\frac{\mathbf{Q}_{1,i}\mathbf{K}_{1,i}^\top}{\sqrt{d_h/2}}\Bigr)\\[1mm]
&\quad - \lambda\,\mathrm{softmax}\!\Bigl(\frac{\mathbf{Q}_{2,i}\mathbf{K}_{2,i}^\top}{\sqrt{d_h/2}}\Bigr)
\Biggr)
\,\mathbf{V}_i,
\end{aligned}
\]
where \(\mathbf{Q}_{1,i}, \mathbf{Q}_{2,i}, \mathbf{K}_{1,i}, \mathbf{K}_{2,i} \in \mathbb{R}^{N \times (d_h/2)}\). The final output is then
\[
\mathrm{DiffMHA}(X)
=
\bigl[
\mathrm{DiffAttn}_1(X)\,\|\,\dots\,\|\,\mathrm{DiffAttn}_h(X)
\bigr]\,
W^O.
\]
By learning complementary attention maps in each head and subtracting them, Diff MHA aims to amplify relevant patterns while reducing noise.

\subsection{CLIP Training}
\label{subsec:clip_pretraining}

Contrastive Language-Image Pre-training (CLIP) \cite{radford2021learning} learns image and text embeddings in a shared space using a large collection of paired image-text examples \(\{(I_k, T_k)\}_{k=1}^{M}\). It consists of two encoders: one for images \(\bigl(f_\theta\bigr)\) and one for text \(\bigl(g_\phi\bigr)\). Their outputs are normalized to unit length:
\[
u_i = \frac{f_\theta(I_i)}{\|f_\theta(I_i)\|_2},
\quad
v_i = \frac{g_\phi(T_i)}{\|g_\phi(T_i)\|_2}.
\]
For a batch of \(N\) pairs, CLIP forms a similarity matrix
\[
S_{ij}
=
\dfrac{u_i^\top \, v_j}{\tau},
\]
where \(\tau\) is a (learned or fixed) temperature parameter. The text-to-image contrastive loss is
\[
\mathcal{L}_{ti}
=
-\tfrac{1}{N}
\sum_{i=1}^{N}
\log
\frac{\exp(S_{ii})}{\sum_{j=1}^{N}\exp(S_{ij})},
\]
and the image-to-text counterpart is
\[
\mathcal{L}_{it}
=
-\frac{1}{N}
\sum_{i=1}^{N}
\log
\tfrac{\exp(S_{ii})}{\sum_{j=1}^{N}\exp(S_{ji})}.
\]
The overall objective is
\[
\mathcal{L}_{\mathrm{CLIP}}
=
\tfrac{1}{2}
\Bigl(\mathcal{L}_{ti}
+
\mathcal{L}_{it}\Bigr).
\]
By encouraging matching image-text pairs to have high similarity (and non-matching pairs to have low similarity), CLIP learns robust features that often transfer well to downstream tasks like zero-shot classification and retrieval.

\begin{table*}[t!]
\centering
\caption{\textbf{Classification Performance (Linear Probing and Few-Shot).}
We compare CLIP and DiffCLIP on nine classification tasks with two pretraining sets (CC3M and CC12M). The top block reports linear probing accuracy, while the bottom block shows few-shot results. Numbers in parentheses indicate absolute gains or drops for DiffCLIP relative to CLIP.}
\label{tab:combined_linear_fewshot}
{\renewcommand{\arraystretch}{1.2} %
\resizebox{\textwidth}{!}{%
\begin{tabular}{llccccccccc|c}
\toprule
\textbf{Pretraining} & \textbf{Model} 
& \textbf{Caltech-101} & \textbf{DTD} & \textbf{Pets} & \textbf{Flowers} & \textbf{SUN397} & \textbf{Aircraft} & \textbf{CIFAR10} & \textbf{CIFAR100} & \textbf{Food-101} & \textbf{Avg.} \\
\midrule
\multicolumn{11}{l}{\textbf{Linear Probing}} \\
\midrule
CC3M & CLIP & 
72.5 & 58.7 & 61.0 & 85.8 & 54.1 & 35.7 & 83.5 & 63.4 & 59.1 & 63.8 \\
CC3M & DiffCLIP & 
76.2\,\posdiff{3.7} & 60.2\,\posdiff{1.5} & 62.2\,\posdiff{1.2} & 86.6\,\posdiff{0.8} & 56.2\,\posdiff{2.1} & 34.6\,\negdiff{1.1} & 83.9\,\posdiff{0.4} & 63.7\,\posdiff{0.3} & 59.4\,\posdiff{0.3} & 64.8\,\posdiff{1.0} \\
\cmidrule(lr){1-12}
CC12M & CLIP & 
88.3 & 71.2 & 79.5 & 92.6 & 68.3 & 48.8 & 92.0 & 74.7 & 77.5 & 77.0 \\
CC12M & DiffCLIP & 
89.5\,\posdiff{1.2} & 71.8\,\posdiff{0.6} & 83.0\,\posdiff{3.5} & 93.5\,\posdiff{0.9} & 69.4\,\posdiff{1.1} & 46.4\,\negdiff{2.4} & 90.7\,\negdiff{1.3} & 73.3\,\negdiff{1.4} & 77.7\,\posdiff{0.2} & 77.3\,\posdiff{0.3} \\
\midrule
\multicolumn{11}{l}{\textbf{Few-Shot}} \\
\midrule
CC3M & CLIP & 
90.4 & 72.9 & 69.6 & 92.5 & 91.8 & 44.6 & 63.4 & 72.8 & 67.0 & 73.9 \\
CC3M & DiffCLIP &
91.6\,\posdiff{1.2} & 73.2\,\posdiff{0.3} & 71.6\,\posdiff{2.0} & 92.9\,\posdiff{0.4} & 92.8\,\posdiff{1.0} & 45.4\,\posdiff{0.8} & 62.4\,\negdiff{1.0} & 73.5\,\posdiff{0.7} & 68.3\,\posdiff{1.3} & 74.6\,\posdiff{0.7} \\
\cmidrule(lr){1-12}
CC12M & CLIP &
97.4 & 81.9 & 86.3 & 96.9 & 96.5 & 56.1 & 81.3 & 85.1 & 86.0 & 85.3 \\
CC12M & DiffCLIP & 
97.6\,\posdiff{0.2} & 82.2\,\posdiff{0.3} & 88.2\,\posdiff{1.9} & 97.3\,\posdiff{0.4} & 96.8\,\posdiff{0.3} & 55.2\,\negdiff{0.9} & 80.3\,\negdiff{1.0} & 83.3\,\negdiff{1.8} & 87.5\,\posdiff{1.5} & 85.4\,\posdiff{0.1} \\
\bottomrule
\end{tabular}%
}}
\end{table*}

\begin{table*}[t!]
\centering
\caption{\textbf{Zero-Shot Retrieval and ImageNet Zero-shot Accuracy.}
We report image and text retrieval (Recall@5, \%) and zero-shot ImageNet accuracy (\%) for CLIP vs.\ DiffCLIP, using CC3M or CC12M as pretraining data. Values in parentheses reflect absolute gains or drops for DiffCLIP relative to CLIP.}
\label{tab:combined_retrieval}
{\renewcommand{\arraystretch}{1.2} %
\resizebox{1.0\textwidth}{!}{%
\begin{tabular}{llcccc|cccc|c}
\toprule
\textbf{Pretraining} & \textbf{Model} 
& \multicolumn{4}{c|}{\textbf{Image Retrieval (R@5)}} 
& \multicolumn{4}{c|}{\textbf{Text Retrieval (R@5)}} 
& \textbf{Zero-Shot} \\
\cmidrule(lr){3-6} \cmidrule(lr){7-10} \cmidrule(lr){11-11}
& 
& \textbf{Flickr30k} & \textbf{Flickr8k} & \textbf{MSCOCO} & \textbf{Avg.} 
& \textbf{Flickr30k} & \textbf{Flickr8k} & \textbf{MSCOCO} & \textbf{Avg.} 
& ImageNet \\
\midrule
CC3M & CLIP & 
31.8 & 35.4 & 19.4 & 28.9 &
43.4 & 46.2 & 25.4 & 38.3 & 13.6 \\
CC3M & DiffCLIP &
32.9\,\posdiff{1.1} & 36.5\,\posdiff{1.1} & 20.9\,\posdiff{1.5} & 30.1\,\posdiff{1.2} &
44.7\,\posdiff{1.3} & 47.8\,\posdiff{1.6} & 27.6\,\posdiff{2.2} & 40.1\,\posdiff{1.8} & 14.4\,\posdiff{0.8} \\
\cmidrule(lr){1-11}
CC12M & CLIP & 
62.5 & 62.1 & 41.3 & 55.3 &
76.8 & 77.7 & 53.8 & 69.4 & 31.8 \\
CC12M & DiffCLIP & 
62.2\,\negdiff{0.3} & 61.5\,\negdiff{0.6} & 42.3\,\posdiff{1.0} & 55.3\,\posdiff{0.0} &
77.4\,\posdiff{0.6} & 77.4\,\negdiff{0.3} & 55.5\,\posdiff{1.7} & 70.1\,\posdiff{0.7} & 33.8\,\posdiff{2.0} \\
\bottomrule
\end{tabular}%
}}
\end{table*}

\section{Experiments}
\label{sec:experiments}

We present an extensive empirical study to investigate whether differential attention can benefit CLIP-style vision-language models. We first describe our dataset sources and training configurations, then evaluate both standard CLIP and our DiffCLIP variant under linear probing, few-shot classification, and image-text retrieval. We also test robustness to distribution shifts (via OOD ImageNet) and fine-grained features (via MMVP), and conclude with ablation studies on the initialization of the differential attention parameter \(\lambda_{\mathrm{init}}\) and on applying differential attention to only the vision encoder.

\subsection{Experimental Setup}
\label{subsec:setup}

\paragraph{Datasets.}
We pretrain on Conceptual Captions 3M (CC3M) \cite{cc3m} and Conceptual Captions 12M (CC12M) \cite{cc12m}. After downloading using \texttt{img2dataset} \cite{beaumont-2021-img2dataset} (with shorter edge resized to 224), we end up with about 2.3M image-text pairs for CC3M and 7.9M for CC12M. For CC3M, we train on four A100 GPUs, while CC12M uses eight A100 GPUs to reduce training time. Text data is minimally processed, limited to basic tokenization.\\

\noindent\textbf{Training Parameters.}\quad
All models train for 40 epochs, using one epoch of linear warmup, a global batch size of 4096, and AdamW optimizer \cite{loshchilov2017decoupled}. We set the base learning rate to \(5 \times 10^{-4}\) with weight decay of 0.5. For DiffCLIP, every attention layer in both the vision and text encoders is replaced with differential attention. We initialize each layer’s \(\lambda\) at 0.8 unless stated otherwise. This setup introduces only a minor parameter overhead: roughly \(0.003\%\) additional parameters relative to a standard CLIP-B/16. Training parameters are chosen similar to SynthCLIP \cite{hammoud2024synthclip} and training code is adopted from SLIP \cite{mu2022slip}.

\paragraph{Evaluation Protocol.}
We follow established practices for linear probing and few-shot evaluation \cite{elbanani2022languageguided} on nine image-classification datasets: DTD~\cite{dtd}, Flowers~\cite{flowers102}, Pets, Caltech-101~\cite{caltech101}, Aircraft~\cite{aircraft}, CIFAR-10~\cite{cifar}, SUN397~\cite{sun397}, CIFAR-100~\cite{cifar}, and Food-101~\cite{food101}. For retrieval (image-to-text and text-to-image) on Flickr8k ~\cite{flickr8k}, Flickr30k~\cite{flickr30k}, and MSCOCO~\cite{mscoco}, we use the LAION CLIP Benchmark framework \cite{schuhmann2022laion}. We measure zero-shot robustness on ImageNet \cite{imagenet} and its variants (ImageNet-V2~\cite{imagenetv2}, ImageNet-A~\cite{imagenetA}, ImageNet-R~\cite{imagenetR}, and ImageNet-Sketch~\cite{imagenetsketch}). Finally, we use the MMVP-VLM benchmark \cite{tong2024eyes} to check how well each model focuses on fine-grained visual details.

\begin{figure*}[t]
    \centering
    \includegraphics[width=\linewidth]{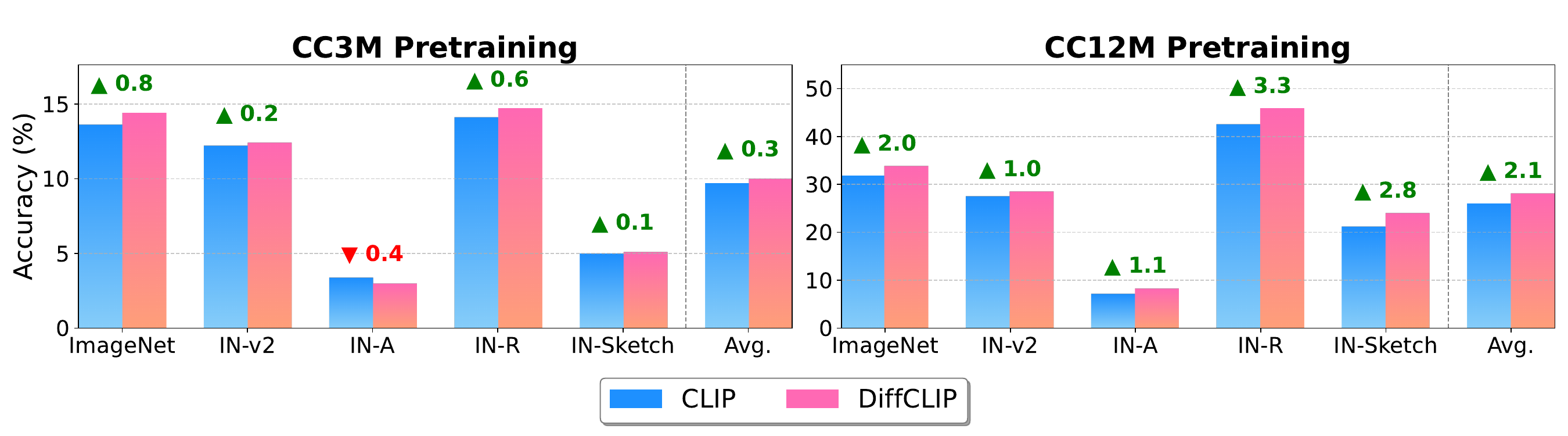}
\caption{\textbf{OOD Zero-Shot ImageNet Performance.}
Comparison of zero-shot accuracy (\%) on ImageNet, ImageNet-V2, ImageNet-A, ImageNet-R, and ImageNet-Sketch, plus the average. Bars show performance of CLIP (blue) versus DiffCLIP (pink), trained on CC3M (left) or CC12M (right). Numerical deltas above the bars indicate the absolute improvement or drop for DiffCLIP relative to CLIP. DiffCLIP improves on average the zero-shot performance on OOD ImageNet datasets as compared to CLIP.}
    \label{fig:ood_imagenet}
\end{figure*}

\subsection{Do CLIP Models Benefit from Differential Attention?}
\textbf{Motivation.}\quad To evaluate the effectiveness of our proposed DiffCLIP, we test its performance across tasks involving image classification, image-text retrieval, and zero-shot generalization, following common benchmarks established in prior literature \cite{hammoud2024synthclip}.

\paragraph{Results.}
We compare a baseline CLIP-B/16 to our DiffCLIP-B/16 (with differential attention in both vision and text encoders). Table~\ref{tab:combined_linear_fewshot} shows linear probing and few-shot classification results for models pretrained on both CC3M and CC12M. DiffCLIP outperforms standard CLIP on almost every dataset. For example, with CC3M pretraining, DiffCLIP achieves about \(\,+1\%\) gain in linear probing and \(\,+0.7\%\) in few-shot accuracy. 

Table~\ref{tab:combined_retrieval} presents retrieval metrics and zero-shot ImageNet. DiffCLIP again surpasses CLIP on image and text retrieval: for CC3M, we see an average improvement of about \(1.2\%\) (image retrieval) and \(1.8\%\) (text retrieval). On zero-shot ImageNet, DiffCLIP-CC3M increases accuracy by \(0.8\%\), with even larger gains of \(\,+2.0\%\) when using CC12M.

\begin{tcolorbox}[width=\linewidth,colback=blue!2,colframe=blue!50!black,arc=2mm,boxrule=0pt,top=10pt,  bottom=10pt]
\textbf{Conclusion.} Even though DiffCLIP only adds a tiny fraction of extra parameters, it consistently outperforms standard CLIP on classification and retrieval benchmarks. This suggests that differential attention is a lightweight yet effective way to enhance vision-language representation.
\end{tcolorbox}

\subsection{Does Differential Attention Improve Out-of-Domain Robustness?}
\label{subsec:ood_improvement}

\paragraph{Motivation.}
Having observed improvements on in-distribution ImageNet, we ask if these gains carry over to more challenging out-of-domain variants. Real-world applications often involve domain shifts, and CLIP’s zero-shot adaptability has been tested on ImageNet-V2, ImageNet-A, ImageNet-R, and ImageNet-Sketch—benchmarks known to stress model robustness beyond standard ImageNet. Understanding how differential attention influences robustness in such scenarios is crucial for assessing its practical utility in deployment settings. We aim to see if differential attention helps maintain or improve performance under such shifts.

\paragraph{Results.}
Figure~\ref{fig:ood_imagenet} summarizes zero-shot performance across ImageNet-V2, ImageNet-A, ImageNet-R, and ImageNet-Sketch. Models with differential attention outperform standard CLIP by an average of \(2.1\%\), suggesting that subtracting noisy attention patterns yields features that generalize more robustly, even under significant distribution shifts.

\begin{tcolorbox}[width=\linewidth,colback=blue!2,colframe=blue!50!black,arc=2mm,boxrule=0pt,top=10pt,  bottom=10pt]
\textbf{Conclusion.} DiffCLIP not only enhances in-distribution performance but also strengthens zero-shot robustness against substantial domain shifts, further demonstrating the benefits of differential attention.
\end{tcolorbox}

\subsection{Does DiffCLIP Improve Fine-Grained Visual Understanding?}
\label{subsec:mmvp}

\vspace{5pt}
\noindent
\textbf{MMVP-VLM Benchmark.}\quad
To test fine-grained visual understanding, we employ the MMVP-VLM benchmark \cite{tong2024eyes}. This benchmark measures how well vision-language models capture nuanced visual properties, such as object orientation, presence, and relational context, beyond straightforward recognition. Both CLIP and DiffCLIP are pretrained on CC12M under identical settings.

\begin{figure}[t]
    \centering
    \includegraphics[width=1.0\linewidth]{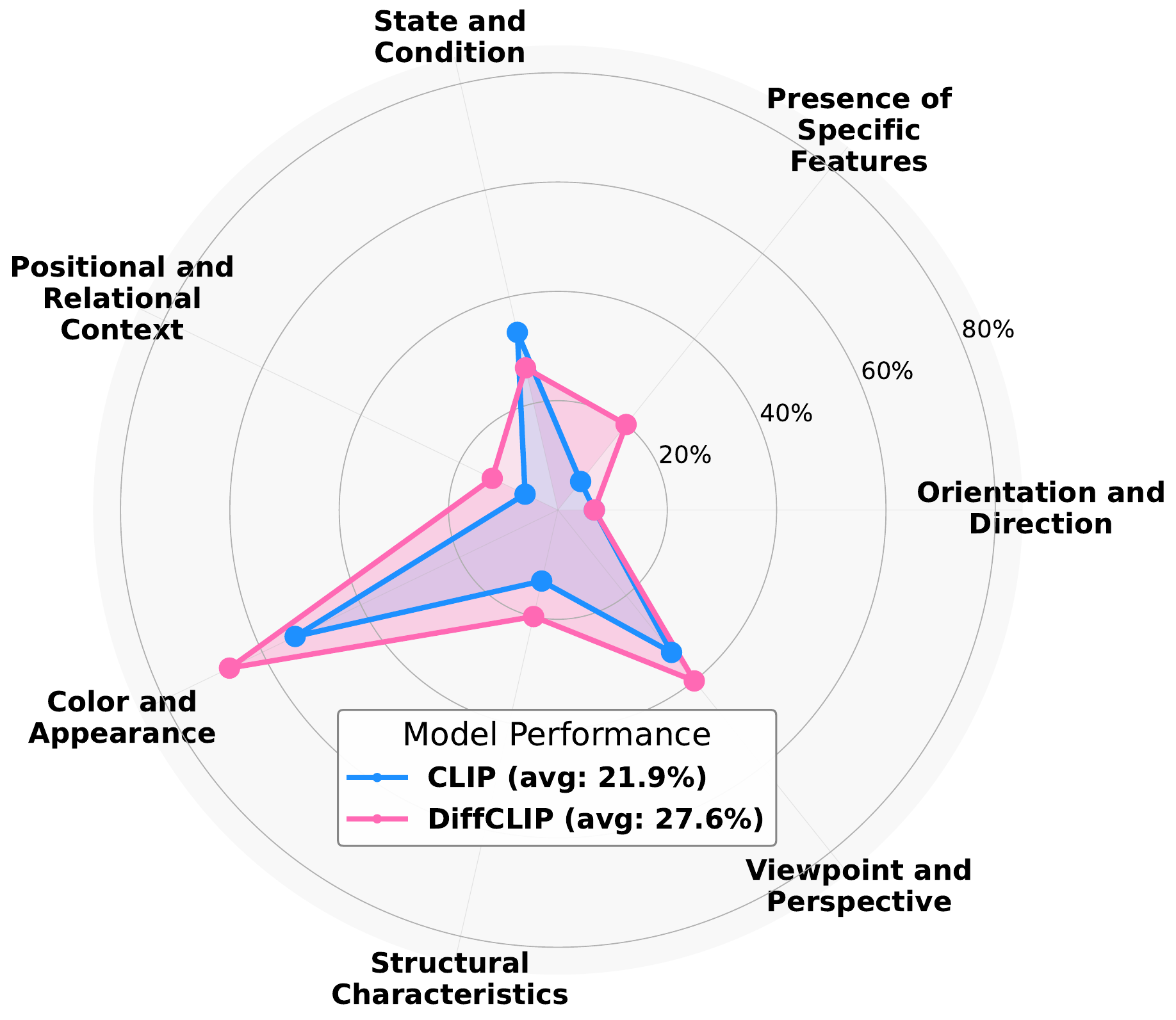}
\caption{\textbf{MMVP-VLM Benchmarking.}
Radar plot illustrating performance on different fine-grained visual categories. Both models (CLIP in blue, DiffCLIP in pink) are evaluated on properties like orientation, positional context, and color appearance. DiffCLIP (average 27.6\%) consistently outperforms CLIP (average 21.9\%), demonstrating more focused attention on subtle visual details.}
    \label{fig:radar_plot}
\end{figure}

\noindent
\textbf{Results.}\quad
On average, DiffCLIP improves MMVP-VLM accuracy by 5.7\% relative to baseline CLIP. A radar plot (Figure~\ref{fig:radar_plot}) shows DiffCLIP surpassing or matching CLIP on nearly all categories except one (\emph{state and condition}). This suggests that subtracting noisy attention patterns (via differential attention) helps the model attend to more subtle details in images.

\begin{tcolorbox}[width=\linewidth,colback=blue!2,colframe=blue!50!black,arc=2mm,boxrule=0pt,top=10pt,  bottom=10pt]
\textbf{Conclusion.} By mitigating extraneous context through differential attention, DiffCLIP achieves stronger fine-grained visual understanding. These gains highlight the effectiveness of explicitly canceling irrelevant attention weights in multimodal settings.
\end{tcolorbox}

\subsection{Dynamic or Static \texorpdfstring{\(\lambda_{\mathrm{init}}\)}{lambda\_init}?}
\label{subsec:lambda_ablation}

\vspace{5pt}
\noindent
\textbf{Motivation.}\quad
All previous experiments used a fixed initialization \(\lambda_{\mathrm{init}} = 0.8\) for differential attention. However, \cite{ye2024differential} proposes a \emph{dynamic} schedule:
\[
\lambda_{\mathrm{init}}(l)
=
0.8
-
0.6
\,\exp(-0.3\,l),
\]
where \(l\) is the layer index. We denote the model using this schedule as \(\text{DiffCLIP}^*\).

\begin{figure*}
    \centering
    \includegraphics[width=1.0\linewidth]{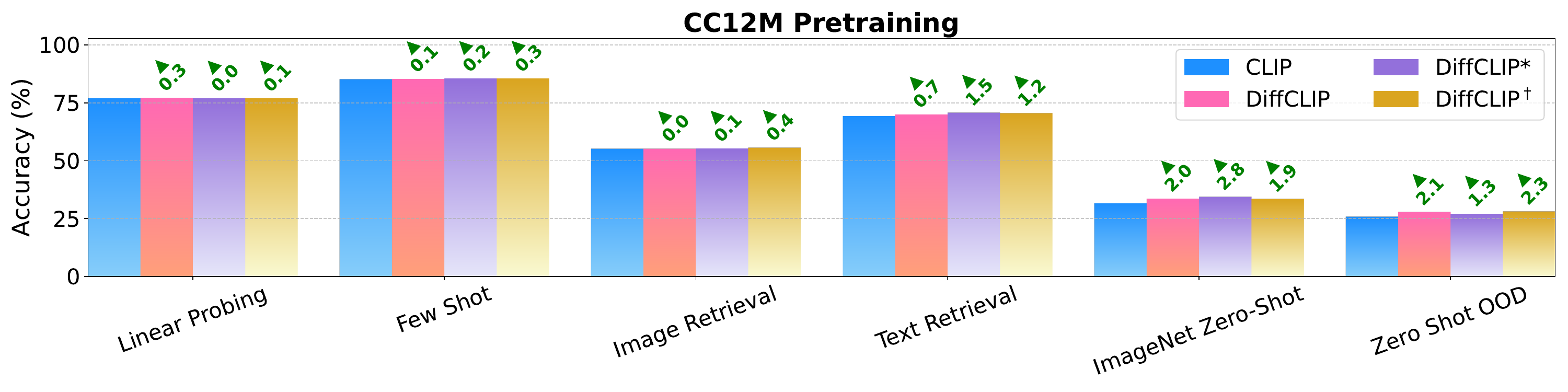}
\caption{\textbf{Comparing Different DiffCLIP Variants.}
We evaluate four models on six tasks (linear probing, few-shot, image retrieval, text retrieval, ImageNet zero-shot, and zero-shot OOD), all pretrained on CC12M. 
CLIP (blue) is the baseline, DiffCLIP (pink) uses a fixed differential attention parameter, 
DiffCLIP$^*$ (purple) employs a dynamic schedule for differential attention, 
and DiffCLIP$^\dagger$ (yellow) applies differential attention only to the vision encoder. }
    \label{fig:ablations_dagger_star}
\end{figure*}

\paragraph{Results.}
Figure~\ref{fig:ablations_dagger_star} summarizes six tasks: linear probing, few-shot classification, image retrieval, text retrieval, zero-shot ImageNet, and zero-shot OOD. Compared to the baseline CC12M CLIP, \(\text{DiffCLIP}^*\) improves zero-shot ImageNet by +2.8\% and text retrieval by +1.5\%. It also raises zero-shot OOD accuracy by +1.3\%. However, relative to \emph{standard} DiffCLIP (with fixed \(\lambda_{\mathrm{init}} = 0.8\)), \(\text{DiffCLIP}^*\) is +0.8\% better on zero-shot ImageNet and +0.8\% on text retrieval, but it \emph{underperforms or only slightly improves} on other tasks. For instance, in zero-shot OOD, \(\text{DiffCLIP}^*\) is -0.8\% behind standard DiffCLIP.

\begin{tcolorbox}[width=\linewidth,colback=blue!2,colframe=blue!50!black,arc=2mm,boxrule=0pt,top=10pt,  bottom=10pt]
\textbf{Conclusion.} A dynamic \(\lambda\) schedule yields notable gains on zero-shot ImageNet and text retrieval, though it lags behind the simpler constant initialization on several other benchmarks. Future work might explore how best to tune or combine these schedules to achieve consistent improvements.
\end{tcolorbox}

\subsection{Does Applying Differential Attention to Vision Only Suffice?}
\label{subsec:vision_only}

\vspace{5pt}
\noindent
\textbf{Motivation.}\quad
Because the vision encoder often plays a dominant role in CLIP models, one might ask whether differential attention is \emph{necessary} in both encoders. We define \(\text{DiffCLIP}^\dagger\) as a variant that integrates differential attention \emph{only in the vision} encoder, leaving the text encoder with regular attention.

\paragraph{Results.}
Figure~\ref{fig:ablations_dagger_star} compares CLIP, DiffCLIP, and \(\text{DiffCLIP}^\dagger\) across six tasks: linear probing, few-shot classification, image retrieval, text retrieval, zero-shot ImageNet, and zero-shot OOD. \(\text{DiffCLIP}^\dagger\) improves upon baseline CLIP by +0.1\% in linear probing, +0.3\% in few-shot, +0.4\% in image retrieval, +1.2\% in text retrieval, +1.9\% on zero-shot ImageNet, and +2.3\% on zero-shot OOD. Compared to  DiffCLIP, \(\text{DiffCLIP}^\dagger\) surpasses or matches performance on few-shot, image retrieval, text retrieval, and zero-shot OOD, but is slightly behind on linear probing and standard zero-shot ImageNet.

\begin{tcolorbox}[width=\linewidth,colback=blue!2,colframe=blue!50!black,arc=2mm,boxrule=0pt,top=10pt,  bottom=10pt]
\textbf{Conclusion.} Applying differential attention solely to the vision encoder already brings sizable gains. Interestingly, \(\text{DiffCLIP}^\dagger\) can even match or exceed full DiffCLIP on certain tasks, suggesting that most of the performance boost may come from more robust visual feature extraction.
\end{tcolorbox}

\begin{table*}[t!]
\centering
\caption{\textbf{CLIP vs. DiffCLIP on POPE Hallucination Benchmark.}
We compare models across three POPE categories, showing accuracy, precision, and recall. Absolute improvements of DiffCLIP over CLIP are highlighted in parentheses.}
\label{tab:clip_diffclip}
\resizebox{0.75\textwidth}{!}{%
\begin{tabular}{lccc|ccc}
\toprule
 & \multicolumn{3}{c|}{\textbf{CLIP}} & \multicolumn{3}{c}{\textbf{DiffCLIP}} \\
\cmidrule(lr){2-4} \cmidrule(lr){5-7}
\textbf{POPE} & Accuracy & Precision & Recall & Accuracy & Precision & Recall \\
\midrule
\textbf{Random} & 
50.14 & 50.07 & 98.21 & 
50.41\,\posdiff{0.27} & 50.21\,\posdiff{0.14} & 98.56\,\posdiff{0.35} \\

\textbf{Popular} & 
50.27 & 50.13 & 99.33 & 
50.27\,\posdiff{0.00} & 50.13\,\posdiff{0.00} & 99.47\,\posdiff{0.14} \\

\textbf{Adversarial} & 
50.17 & 50.08 & 99.33 & 
50.20\,\posdiff{0.03} & 50.10\,\posdiff{0.02} & 99.47\,\posdiff{0.14} \\

\bottomrule
\end{tabular}
}
\end{table*}

\section{Future Directions \& Limitations}\label{sec:future}
\subsection{Beyond CLIP}

An intriguing question for future research is how a vision encoder trained with differential attention within the CLIP framework would perform when integrated into larger, more sophisticated vision-language models such as LLaVA \cite{liu2023visual} or TinyLLaVA \cite{zhou2024tinyllava}. To provide initial insights into this possibility, we conducted preliminary experiments by combining our DiffCLIP-CC12M vision encoder with the Qwen-2.5-Instruct-0.5B \cite{yang2024qwen2} language encoder. We followed a typical two-stage training procedure: first, a linear projector was trained to align visual tokens with the language embedding space, freezing all other components; second, both the projector and the language encoder underwent instruction fine-tuning.

For the projection training, we utilized the LAION-CC-SBU dataset (558K image-text pairs) used in the LLaVA training setup. For instruction fine-tuning, we adopted the COCO \cite{mscoco} subset (approximately 350K pairs) also used by LLaVA. All experiments were conducted using the TinyLLaVA repository on 4 A100-80GB GPUs. The hyperparameters for fine-tuning included a batch size of 48 samples per GPU, a learning rate of $2 \times 10^{-5}$, zero weight decay, a warm-up ratio of 0.03, and cosine decay scheduling. The projection pretraining similarly employed 48 samples per GPU, a learning rate of $1 \times 10^{-3}$, no weight decay, a warm-up ratio of 0.03, and cosine decay scheduling.

We evaluated the resulting models on the POPE \cite{pope} hallucination dataset, which assesses models' susceptibility to visual hallucinations. Despite the modest size of observed improvements, DiffCLIP-CC12M consistently outperformed the CLIP-CC12M baseline across all metrics. These initial findings suggest that differential attention-trained vision encoders could enhance performance when integrated into broader vision-language architectures, making it a promising direction for further exploration.

\subsection{Scaling Data and Architecture}

Training CLIP models with a ViT-B/16 backbone on the CC12M dataset (7.9M samples) currently requires approximately 10 GPU-days on A100 GPUs, translating to roughly \$600 using Google Cloud Platform (GCP). A natural future direction would involve exploring how differential attention performs when scaling to larger architectures (e.g., ViT-L or ViT-H) and substantially bigger datasets (e.g., LAION-400M). Investigating such scaling could reveal whether the performance gains observed with DiffCLIP persist or even amplify as model size and dataset scale increase, offering insights into the broader applicability and benefits of differential attention in vision-language pretraining.
\section{Conclusion}
\label{sec:conclusion}

We introduced DiffCLIP, which integrates differential attention into CLIP-based vision-language models to better filter out noisy alignments. Through extensive experiments on classification, retrieval, robustness, and fine-grained benchmarks, DiffCLIP consistently improves over standard CLIP with minimal overhead. Further ablations highlight the flexibility of dynamic attention schedules and vision-only setups. We hope these findings inspire future research on more efficient, robust attention mechanisms in large multimodal learning.

\section{Acknowledgements}

The research reported in this publication was supported by funding from King Abdullah University of Science and Technology (KAUST) - Center of Excellence for Generative AI, under award number 5940.
{
    \small
    \bibliographystyle{ieeenat_fullname}
    \bibliography{main}

\begin{thebibliography}{54}
\providecommand{\natexlab}[1]{#1}
\providecommand{\url}[1]{\texttt{#1}}
\expandafter\ifx\csname urlstyle\endcsname\relax
  \providecommand{\doi}[1]{doi: #1}\else
  \providecommand{\doi}{doi: \begingroup \urlstyle{rm}\Url}\fi

\bibitem[Abbas et~al.(2023)Abbas, Tirumala, Simig, Ganguli, and Morcos]{abbas2023semdedup}
Amro Abbas, Kushal Tirumala, Dániel Simig, Surya Ganguli, and Ari~S. Morcos.
\newblock Semdedup: Data-efficient learning at web-scale through semantic deduplication, 2023.

\bibitem[Beaumont(2021)]{beaumont-2021-img2dataset}
Romain Beaumont.
\newblock img2dataset: Easily turn large sets of image urls to an image dataset.
\newblock \url{https://github.com/rom1504/img2dataset}, 2021.

\bibitem[Bossard et~al.(2014)Bossard, Guillaumin, and Van~Gool]{food101}
Lukas Bossard, Matthieu Guillaumin, and Luc Van~Gool.
\newblock Food-101--mining discriminative components with random forests.
\newblock In \emph{ECCV}, pages 446--461. Springer, 2014.

\bibitem[Changpinyo et~al.(2021)Changpinyo, Sharma, Ding, and Soricut]{cc12m}
Soravit Changpinyo, Piyush Sharma, Nan Ding, and Radu Soricut.
\newblock {Conceptual 12M}: Pushing web-scale image-text pre-training to recognize long-tail visual concepts.
\newblock In \emph{CVPR}, 2021.

\bibitem[Chen et~al.(2024)Chen, Yu, Shen, Yuille, and Chen]{chen2024vitamin}
Jieneng Chen, Qihang Yu, Xiaohui Shen, Alan Yuille, and Liang-Chieh Chen.
\newblock Vitamin: Designing scalable vision models in the vision-language era.
\newblock In \emph{CVPR}, pages 12954--12966, 2024.

\bibitem[Cimpoi et~al.(2014)Cimpoi, Maji, Kokkinos, Mohamed, and Vedaldi]{dtd}
M. Cimpoi, S. Maji, I. Kokkinos, S. Mohamed, and A. Vedaldi.
\newblock Describing textures in the wild.
\newblock In \emph{CVPR}, 2014.

\bibitem[Dosovitskiy et~al.(2020)Dosovitskiy, Beyer, Kolesnikov, Weissenborn, Zhai, Unterthiner, Dehghani, Minderer, Heigold, Gelly, et~al.]{dosovitskiy2020image}
Alexey Dosovitskiy, Lucas Beyer, Alexander Kolesnikov, Dirk Weissenborn, Xiaohua Zhai, Thomas Unterthiner, Mostafa Dehghani, Matthias Minderer, Georg Heigold, Sylvain Gelly, et~al.
\newblock An image is worth 16x16 words: Transformers for image recognition at scale.
\newblock \emph{arXiv preprint arXiv:2010.11929}, 2020.

\bibitem[El~Banani et~al.(2023)El~Banani, Desai, and Johnson]{elbanani2022languageguided}
Mohamed El~Banani, Karan Desai, and Justin Johnson.
\newblock {Learning Visual Representations via Language-Guided Sampling}.
\newblock In \emph{CVPR}, 2023.

\bibitem[Gadre et~al.(2023)Gadre, Ilharco, Fang, Hayase, Smyrnis, Nguyen, Marten, Wortsman, Ghosh, Zhang, Orgad, Entezari, Daras, Pratt, Ramanujan, Bitton, Marathe, Mussmann, Vencu, Cherti, Krishna, Koh, Saukh, Ratner, Song, Hajishirzi, Farhadi, Beaumont, Oh, Dimakis, Jitsev, Carmon, Shankar, and Schmidt]{gadre2023datacomp}
Samir~Yitzhak Gadre, Gabriel Ilharco, Alex Fang, Jonathan Hayase, Georgios Smyrnis, Thao Nguyen, Ryan Marten, Mitchell Wortsman, Dhruba Ghosh, Jieyu Zhang, Eyal Orgad, Rahim Entezari, Giannis Daras, Sarah Pratt, Vivek Ramanujan, Yonatan Bitton, Kalyani Marathe, Stephen Mussmann, Richard Vencu, Mehdi Cherti, Ranjay Krishna, Pang~Wei Koh, Olga Saukh, Alexander Ratner, Shuran Song, Hannaneh Hajishirzi, Ali Farhadi, Romain Beaumont, Sewoong Oh, Alex Dimakis, Jenia Jitsev, Yair Carmon, Vaishaal Shankar, and Ludwig Schmidt.
\newblock Datacomp: In search of the next generation of multimodal datasets, 2023.

\bibitem[Gan et~al.(2022)Gan, Li, Li, Wang, Liu, Gao, et~al.]{gan2022vision}
Zhe Gan, Linjie Li, Chunyuan Li, Lijuan Wang, Zicheng Liu, Jianfeng Gao, et~al.
\newblock Vision-language pre-training: Basics, recent advances, and future trends.
\newblock \emph{Foundations and Trends{\textregistered} in Computer Graphics and Vision}, 14\penalty0 (3--4):\penalty0 163--352, 2022.

\bibitem[Gao et~al.(2022)Gao, Liu, Xu, Zhang, Li, Ji, and Shen]{gao2022pyramidclip}
Yuting Gao, Jinfeng Liu, Zihan Xu, Jun Zhang, Ke Li, Rongrong Ji, and Chunhua Shen.
\newblock Pyramidclip: Hierarchical feature alignment for vision-language model pretraining.
\newblock \emph{NeurIPS}, 35:\penalty0 35959--35970, 2022.

\bibitem[Hammoud et~al.(2024)Hammoud, Itani, Pizzati, Torr, Bibi, and Ghanem]{hammoud2024synthclip}
Hasan Abed Al~Kader Hammoud, Hani Itani, Fabio Pizzati, Philip Torr, Adel Bibi, and Bernard Ghanem.
\newblock Synthclip: Are we ready for a fully synthetic clip training?
\newblock \emph{arXiv preprint arXiv:2402.01832}, 2024.

\bibitem[He et~al.(2016)He, Zhang, Ren, and Sun]{he2016deep}
Kaiming He, Xiangyu Zhang, Shaoqing Ren, and Jian Sun.
\newblock Deep residual learning for image recognition.
\newblock In \emph{CVPR}, pages 770--778, 2016.

\bibitem[Hendrycks et~al.(2021{\natexlab{a}})Hendrycks, Basart, Mu, Kadavath, Wang, Dorundo, Desai, Zhu, Parajuli, Guo, et~al.]{imagenetR}
Dan Hendrycks, Steven Basart, Norman Mu, Saurav Kadavath, Frank Wang, Evan Dorundo, Rahul Desai, Tyler Zhu, Samyak Parajuli, Mike Guo, et~al.
\newblock The many faces of robustness: A critical analysis of out-of-distribution generalization.
\newblock In \emph{ICCV}, pages 8340--8349, 2021{\natexlab{a}}.

\bibitem[Hendrycks et~al.(2021{\natexlab{b}})Hendrycks, Zhao, Basart, Steinhardt, and Song]{imagenetA}
Dan Hendrycks, Kevin Zhao, Steven Basart, Jacob Steinhardt, and Dawn Song.
\newblock Natural adversarial examples.
\newblock In \emph{CVPR}, pages 15262--15271, 2021{\natexlab{b}}.

\bibitem[Iscen et~al.(2023)Iscen, Caron, Fathi, and Schmid]{iscen2023retrieval}
Ahmet Iscen, Mathilde Caron, Alireza Fathi, and Cordelia Schmid.
\newblock Retrieval-enhanced contrastive vision-text models.
\newblock \emph{arXiv preprint arXiv:2306.07196}, 2023.

\bibitem[Jia et~al.(2021)Jia, Yang, Xia, Chen, Parekh, Pham, Le, Sung, Li, and Duerig]{jia2021scaling}
Chao Jia, Yinfei Yang, Ye Xia, Yi-Ting Chen, Zarana Parekh, Hieu Pham, Quoc Le, Yun-Hsuan Sung, Zhen Li, and Tom Duerig.
\newblock Scaling up visual and vision-language representation learning with noisy text supervision.
\newblock In \emph{ICML}, pages 4904--4916. PMLR, 2021.

\bibitem[Kamradt(2023)]{needle}
Greg Kamradt.
\newblock Needle in a {Haystack} - pressure testing {LLMs}.
\newblock \url{https://github.com/gkamradt/LLMTest_NeedleInAHaystack/tree/main}, 2023.

\bibitem[Krizhevsky et~al.(2009)Krizhevsky, Hinton, et~al.]{cifar}
Alex Krizhevsky, Geoffrey Hinton, et~al.
\newblock Learning multiple layers of features from tiny images.
\newblock 2009.

\bibitem[Lai et~al.(2024)Lai, Zhang, Zhang, Wu, Bai, Timofeev, Du, Gan, Shan, Chuah, et~al.]{lai2024veclip}
Zhengfeng Lai, Haotian Zhang, Bowen Zhang, Wentao Wu, Haoping Bai, Aleksei Timofeev, Xianzhi Du, Zhe Gan, Jiulong Shan, Chen-Nee Chuah, et~al.
\newblock Veclip: Improving clip training via visual-enriched captions.
\newblock In \emph{ECCV}, pages 111--127. Springer, 2024.

\bibitem[Li et~al.(2022)Li, Andreeto, Ranzato, and Perona]{caltech101}
Fei-Fei Li, Marco Andreeto, Marc'Aurelio Ranzato, and Pietro Perona.
\newblock Caltech 101, 2022.

\bibitem[Li et~al.(2021)Li, Liang, Zhao, Cui, Ouyang, Shao, Yu, and Yan]{li2021supervision}
Yangguang Li, Feng Liang, Lichen Zhao, Yufeng Cui, Wanli Ouyang, Jing Shao, Fengwei Yu, and Junjie Yan.
\newblock Supervision exists everywhere: A data efficient contrastive language-image pre-training paradigm.
\newblock \emph{arXiv preprint arXiv:2110.05208}, 2021.

\bibitem[Li et~al.(2023)Li, Du, Zhou, Wang, Zhao, and Wen]{pope}
Yifan Li, Yifan Du, Kun Zhou, Jinpeng Wang, Xin Zhao, and Ji-Rong Wen.
\newblock Evaluating object hallucination in large vision-language models.
\newblock In \emph{ACL}, pages 292--305, Singapore, 2023. Association for Computational Linguistics.

\bibitem[Lin et~al.(2014)Lin, Maire, Belongie, Hays, Perona, Ramanan, Doll{\'a}r, and Zitnick]{mscoco}
Tsung-Yi Lin, Michael Maire, Serge Belongie, James Hays, Pietro Perona, Deva Ramanan, Piotr Doll{\'a}r, and C~Lawrence Zitnick.
\newblock Microsoft coco: Common objects in context.
\newblock In \emph{ECCV}, pages 740--755. Springer, 2014.

\bibitem[Liu et~al.(2023)Liu, Li, Wu, and Lee]{liu2023visual}
Haotian Liu, Chunyuan Li, Qingyang Wu, and Yong~Jae Lee.
\newblock Visual instruction tuning.
\newblock \emph{NeurIPS}, 36:\penalty0 34892--34916, 2023.

\bibitem[Liu et~al.(2024{\natexlab{a}})Liu, Lin, Hewitt, Paranjape, Bevilacqua, Petroni, and Liang]{lost}
Nelson~F Liu, Kevin Lin, John Hewitt, Ashwin Paranjape, Michele Bevilacqua, Fabio Petroni, and Percy Liang.
\newblock Lost in the middle: How language models use long contexts.
\newblock \emph{Transactions of the Association for Computational Linguistics}, 12:\penalty0 157--173, 2024{\natexlab{a}}.

\bibitem[Liu et~al.(2024{\natexlab{b}})Liu, Li, Wang, Zhao, and Xie]{liu2024clips}
Yanqing Liu, Xianhang Li, Zeyu Wang, Bingchen Zhao, and Cihang Xie.
\newblock Clips: An enhanced clip framework for learning with synthetic captions.
\newblock \emph{arXiv preprint arXiv:2411.16828}, 2024{\natexlab{b}}.

\bibitem[Loshchilov and Hutter(2017)]{loshchilov2017decoupled}
Ilya Loshchilov and Frank Hutter.
\newblock Decoupled weight decay regularization.
\newblock \emph{arXiv preprint arXiv:1711.05101}, 2017.

\bibitem[Luo et~al.(2024)Luo, Shi, Khan, Afzal, Huang, Yuan, Tian, Song, Kouhana, Elze, et~al.]{luo2024fairclip}
Yan Luo, Min Shi, Muhammad~Osama Khan, Muhammad~Muneeb Afzal, Hao Huang, Shuaihang Yuan, Yu Tian, Luo Song, Ava Kouhana, Tobias Elze, et~al.
\newblock Fairclip: Harnessing fairness in vision-language learning.
\newblock In \emph{CVPR}, pages 12289--12301, 2024.

\bibitem[Maji et~al.(2013)Maji, Kannala, Rahtu, Blaschko, and Vedaldi]{aircraft}
S. Maji, J. Kannala, E. Rahtu, M. Blaschko, and A. Vedaldi.
\newblock Fine-grained visual classification of aircraft.
\newblock Technical report, 2013.

\bibitem[Mu et~al.(2022)Mu, Kirillov, Wagner, and Xie]{mu2022slip}
Norman Mu, Alexander Kirillov, David Wagner, and Saining Xie.
\newblock Slip: Self-supervision meets language-image pre-training.
\newblock In \emph{ECCV}, pages 529--544. Springer, 2022.

\bibitem[Nilsback and Zisserman(2008)]{flowers102}
Maria-Elena Nilsback and Andrew Zisserman.
\newblock Automated flower classification over a large number of classes.
\newblock In \emph{2008 Sixth Indian conference on computer vision, graphics \& image processing}, pages 722--729. IEEE, 2008.

\bibitem[Radford et~al.(2021)Radford, Kim, Hallacy, Ramesh, Goh, Agarwal, Sastry, Askell, Mishkin, Clark, et~al.]{radford2021learning}
Alec Radford, Jong~Wook Kim, Chris Hallacy, Aditya Ramesh, Gabriel Goh, Sandhini Agarwal, Girish Sastry, Amanda Askell, Pamela Mishkin, Jack Clark, et~al.
\newblock Learning transferable visual models from natural language supervision.
\newblock In \emph{ICML}, pages 8748--8763. PmLR, 2021.

\bibitem[Rashtchian et~al.(2010)Rashtchian, Young, Hodosh, and Hockenmaier]{flickr8k}
Cyrus Rashtchian, Peter Young, Micah Hodosh, and Julia Hockenmaier.
\newblock Collecting image annotations using amazon’s mechanical turk.
\newblock In \emph{Proceedings of the NAACL HLT 2010 workshop on creating speech and language data with Amazon’s Mechanical Turk}, pages 139--147, 2010.

\bibitem[Recht et~al.(2019)Recht, Roelofs, Schmidt, and Shankar]{imagenetv2}
Benjamin Recht, Rebecca Roelofs, Ludwig Schmidt, and Vaishaal Shankar.
\newblock Do imagenet classifiers generalize to imagenet?
\newblock In \emph{ICML}, pages 5389--5400. PMLR, 2019.

\bibitem[Russakovsky et~al.(2015)Russakovsky, Deng, Su, Krause, Satheesh, Ma, Huang, Karpathy, Khosla, Bernstein, et~al.]{imagenet}
Olga Russakovsky, Jia Deng, Hao Su, Jonathan Krause, Sanjeev Satheesh, Sean Ma, Zhiheng Huang, Andrej Karpathy, Aditya Khosla, Michael Bernstein, et~al.
\newblock Imagenet large scale visual recognition challenge.
\newblock \emph{IJCV}, 115:\penalty0 211--252, 2015.

\bibitem[Schuhmann et~al.(2022)Schuhmann, Beaumont, Vencu, Gordon, Wightman, Cherti, Coombes, Katta, Mullis, Wortsman, et~al.]{schuhmann2022laion}
Christoph Schuhmann, Romain Beaumont, Richard Vencu, Cade Gordon, Ross Wightman, Mehdi Cherti, Theo Coombes, Aarush Katta, Clayton Mullis, Mitchell Wortsman, et~al.
\newblock Laion-5b: An open large-scale dataset for training next generation image-text models.
\newblock \emph{NeurIPS}, 35:\penalty0 25278--25294, 2022.

\bibitem[Sharma et~al.(2018)Sharma, Ding, Goodman, and Soricut]{cc3m}
Piyush Sharma, Nan Ding, Sebastian Goodman, and Radu Soricut.
\newblock Conceptual captions: A cleaned, hypernymed, image alt-text dataset for automatic image captioning.
\newblock In \emph{ACL}, 2018.

\bibitem[Tong et~al.(2024)Tong, Liu, Zhai, Ma, LeCun, and Xie]{tong2024eyes}
Shengbang Tong, Zhuang Liu, Yuexiang Zhai, Yi Ma, Yann LeCun, and Saining Xie.
\newblock Eyes wide shut? exploring the visual shortcomings of multimodal llms.
\newblock In \emph{CVPR}, pages 9568--9578, 2024.

\bibitem[Tschannen et~al.(2022)Tschannen, Mustafa, and Houlsby]{tschannen2022image}
Michael Tschannen, Basil Mustafa, and Neil Houlsby.
\newblock Image-and-language understanding from pixels only.
\newblock \emph{arXiv preprint arXiv:2212.08045}, 2022.

\bibitem[Vaswani et~al.(2017)Vaswani, Shazeer, Parmar, Uszkoreit, Jones, Gomez, Kaiser, and Polosukhin]{vaswani2017attention}
Ashish Vaswani, Noam Shazeer, Niki Parmar, Jakob Uszkoreit, Llion Jones, Aidan~N Gomez, {\L}ukasz Kaiser, and Illia Polosukhin.
\newblock Attention is all you need.
\newblock \emph{NeurIPS}, 30, 2017.

\bibitem[Wang et~al.(2019)Wang, Ge, Lipton, and Xing]{imagenetsketch}
Haohan Wang, Songwei Ge, Zachary Lipton, and Eric~P Xing.
\newblock Learning robust global representations by penalizing local predictive power.
\newblock \emph{NeurIPS}, 32, 2019.

\bibitem[Wei et~al.(2024)Wei, Pan, and Owens]{wei2024efficient}
Zihao Wei, Zixuan Pan, and Andrew Owens.
\newblock Efficient vision-language pre-training by cluster masking.
\newblock In \emph{CVPR}, pages 26815--26825, 2024.

\bibitem[Xiao et~al.(2010)Xiao, Hays, Ehinger, Oliva, and Torralba]{sun397}
Jianxiong Xiao, James Hays, Krista~A Ehinger, Aude Oliva, and Antonio Torralba.
\newblock Sun database: Large-scale scene recognition from abbey to zoo.
\newblock In \emph{CVPR}, pages 3485--3492. IEEE, 2010.

\bibitem[Xu et~al.(2022)Xu, De~Mello, Liu, Byeon, Breuel, Kautz, and Wang]{xu2022groupvit}
Jiarui Xu, Shalini De~Mello, Sifei Liu, Wonmin Byeon, Thomas Breuel, Jan Kautz, and Xiaolong Wang.
\newblock Groupvit: Semantic segmentation emerges from text supervision.
\newblock \emph{arXiv preprint arXiv:2202.11094}, 2022.

\bibitem[Yang et~al.(2024)Yang, Yang, Zhang, Hui, Zheng, Yu, Li, Liu, Huang, Wei, et~al.]{yang2024qwen2}
An Yang, Baosong Yang, Beichen Zhang, Binyuan Hui, Bo Zheng, Bowen Yu, Chengyuan Li, Dayiheng Liu, Fei Huang, Haoran Wei, et~al.
\newblock Qwen2. 5 technical report.
\newblock \emph{arXiv preprint arXiv:2412.15115}, 2024.

\bibitem[Yang et~al.(2022)Yang, Li, Zhang, Xiao, Liu, Yuan, and Gao]{yang2022unified}
Jianwei Yang, Chunyuan Li, Pengchuan Zhang, Bin Xiao, Ce Liu, Lu Yuan, and Jianfeng Gao.
\newblock Unified contrastive learning in image-text-label space.
\newblock In \emph{CVPR}, pages 19163--19173, 2022.

\bibitem[Ye et~al.(2024)Ye, Dong, Xia, Sun, Zhu, Huang, and Wei]{ye2024differential}
Tianzhu Ye, Li Dong, Yuqing Xia, Yutao Sun, Yi Zhu, Gao Huang, and Furu Wei.
\newblock Differential transformer.
\newblock \emph{arXiv preprint arXiv:2410.05258}, 2024.

\bibitem[Young et~al.(2014)Young, Lai, Hodosh, and Hockenmaier]{flickr30k}
Peter Young, Alice Lai, Micah Hodosh, and Julia Hockenmaier.
\newblock From image descriptions to visual denotations: New similarity metrics for semantic inference over event descriptions.
\newblock \emph{Transactions of the Association for Computational Linguistics}, 2:\penalty0 67--78, 2014.

\bibitem[Zhai et~al.(2023)Zhai, Mustafa, Kolesnikov, and Beyer]{zhai2023sigmoid}
Xiaohua Zhai, Basil Mustafa, Alexander Kolesnikov, and Lucas Beyer.
\newblock Sigmoid loss for language image pre-training.
\newblock In \emph{ICCV}, pages 11975--11986, 2023.

\bibitem[Zhang et~al.(2024)Zhang, Huang, Jin, and Lu]{zhang2024vision}
Jingyi Zhang, Jiaxing Huang, Sheng Jin, and Shijian Lu.
\newblock Vision-language models for vision tasks: A survey.
\newblock \emph{IEEE TPAMI}, 2024.

\bibitem[Zheng et~al.(2024)Zheng, Zhang, Kembhavi, and Krishna]{zheng2024iterated}
Chenhao Zheng, Jieyu Zhang, Aniruddha Kembhavi, and Ranjay Krishna.
\newblock Iterated learning improves compositionality in large vision-language models.
\newblock In \emph{CVPR}, pages 13785--13795, 2024.

\bibitem[Zhong et~al.(2022)Zhong, Yang, Zhang, Li, Codella, Li, Zhou, Dai, Yuan, Li, et~al.]{zhong2022regionclip}
Yiwu Zhong, Jianwei Yang, Pengchuan Zhang, Chunyuan Li, Noel Codella, Liunian~Harold Li, Luowei Zhou, Xiyang Dai, Lu Yuan, Yin Li, et~al.
\newblock Regionclip: Region-based language-image pretraining.
\newblock In \emph{CVPR}, pages 16793--16803, 2022.

\bibitem[Zhou et~al.(2024)Zhou, Hu, Weng, Jia, Luo, Liu, Wu, and Huang]{zhou2024tinyllava}
Baichuan Zhou, Ying Hu, Xi Weng, Junlong Jia, Jie Luo, Xien Liu, Ji Wu, and Lei Huang.
\newblock Tinyllava: A framework of small-scale large multimodal models.
\newblock \emph{arXiv preprint arXiv:2402.14289}, 2024.

\end{thebibliography}
}

\end{document}